\documentclass{article} % For LaTeX2e
\usepackage[final]{colm2026_conference}

\usepackage{microtype}
\usepackage{hyperref}
\usepackage{url}
\usepackage{booktabs}

\usepackage{hyperref}
\usepackage{url}
\usepackage{booktabs}
\usepackage{amsfonts,amsmath}
\usepackage{nicefrac}
\usepackage{microtype}
\usepackage{xcolor}
\usepackage{graphicx}
\usepackage{multirow}
\usepackage{makecell}
\usepackage{array}
\usepackage{subcaption}
\usepackage{enumitem}
\usepackage{pgfplots}
\usepackage{pgfplotstable}
\pgfplotsset{compat=1.18}
\usepackage{tikz}
\usetikzlibrary{shapes.geometric,arrows.meta,positioning,calc,patterns}

\definecolor{coachcol}{HTML}{4472C4}
\definecolor{warmcol}{HTML}{ED7D31}
\definecolor{tsukcol}{HTML}{70AD47}
\definecolor{realcol}{HTML}{FF0000}
\definecolor{gonzocol}{HTML}{7030A0}
\definecolor{lightgray}{HTML}{F2F2F2}
\definecolor{accentblue}{HTML}{2E75B6}

\usepackage{lineno}

\definecolor{darkblue}{rgb}{0, 0, 0.5}
\hypersetup{colorlinks=true, citecolor=darkblue, linkcolor=darkblue, urlcolor=darkblue}

\newcommand{\sys}{\textsc{Ekova}}
\newcommand{\backbone}{\textsc{DeepSupport}}

\newcommand{\Psup}{Personality Support}

\newcommand{\dsd}{\textit{DSD}}
\newcommand{\orthotune}{\textit{OrthoTune}}

\newcommand{\dataname}{\raisebox{-0.05\height}{\includegraphics[width=22pt]{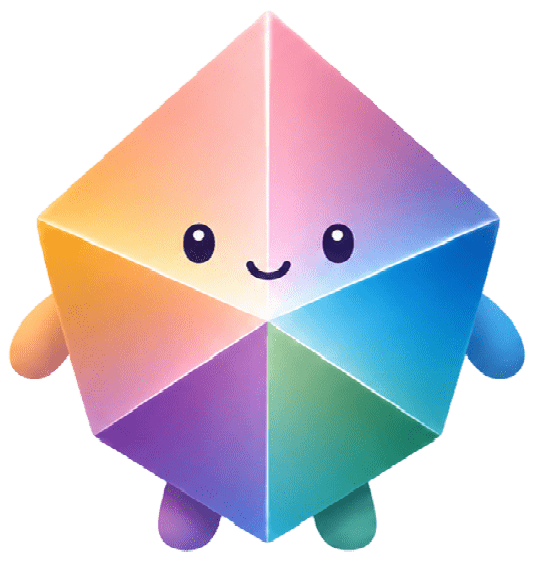}}\textsc{Ekova}}

\title{\dataname{}: A Personality-Support Agent for Self-Discovery Dialogue}

\author{
Yuyan Chen\thanks{Yuyan Chen is the Founder and CEO of \href{http://modelslive.org/}{ModelsLive Inc.}} \\
ModelsLive Inc. \\
\texttt{ychen@modelslive.org}
}

\begin{document}

\ifcolmsubmission
\linenumbers
\fi

\maketitle

\begin{abstract}
Emotional Support (ES) systems have long optimized a single objective: alleviating the user's emotional distress in the moment. We argue that a complementary need, helping users \emph{see themselves more clearly}, defines a distinct paradigm we call \Psup{} (PS). PS is not counseling or clinical intervention: it targets cognitive clarity and self-articulation, not symptom relief or diagnosis.
We instantiate this paradigm in three layers. First, we present \dsd{}, a Chinese self-discovery PS Dataset of 8,590 samples collected through real longitudinal interaction across five minimal units, Coach, Warm, Tsukkomi, Real, and Gonzo. Second, we build \backbone{}, a multi-persona PS system trained with \orthotune{}, a PS-tailored framework with style-specific adapters and a style-consistency regularizer. Third, we unify the five \backbone{} personas into \sys{}, a persistent personality-support agent with a unified cross-session memory layer, supporting both adaptive routing and user-customized persona selection. Experiments show that \orthotune{}-trained models outperform all baselines with an average relative gain of 16.3\% across all metrics over the strongest prompt-based baseline. Code is available at \url{https://github.com/Yukyin/Ekova}.
\end{abstract}
% ============================================================
\section{Introduction}
\label{sec:intro}

\begin{figure}[t]
 \centering
 \includegraphics[width=0.95\linewidth]{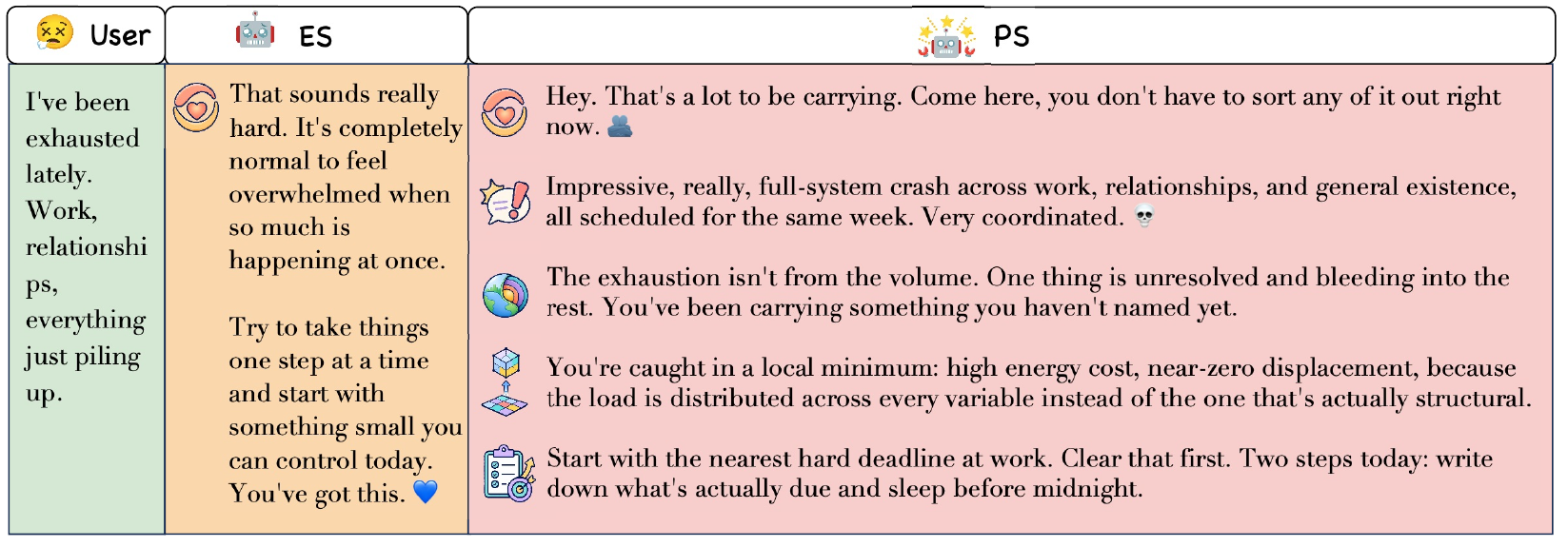}
 \caption{\small
 The same user input processed under Emotional Support (ES) and Personality Support (PS). ES optimizes for affect regulation. PS targets cognitive clarity through grounding, reframing, de-narrativizing, elevating, and directional closure.
 }
 \label{fig:paradigm}
\end{figure}

Emotional Support (ES) systems have made significant progress in helping users navigate personal setbacks and emotional crises, offering empathic responses and evidence-based coping strategies~\citep{liu2021towards, chen2023soulchat, zhang2024cpsycoun}. However, most existing ES systems optimize for a single reactive objective: alleviating the user's emotional distress in the moment. This framing overlooks a complementary and often more enduring need, namely Personality Support (PS): helping users articulate their problems more clearly, discover overlooked self-patterns, and recognize the structural reality of their situation. PS is reflexive rather than reactive, and aims at durable self-understanding rather than momentary affect regulation. Critically, PS is not counseling or clinical intervention: it targets the cognitive layer, not symptom relief or diagnosis.
This paper keeps the construction progression explicit. \dsd{} is the dataset layer, \backbone{} is the trained multi-persona system, and \sys{} is the unified persistent agent that integrates the five \backbone{} personas into one coherent support experience while preserving the underlying PS paradigm.
As Figure~\ref{fig:paradigm} illustrates, the same user input processed under ES yields a single affect-regulation response, while PS 
produces a qualitatively different kind of engagement: it 
grounds the user emotionally, reframes the situation with cognitive distance, de-narrativizes the underlying structure, elevates via cross-domain analogy, and closes with a concrete action plan. ES asks \emph{how does the user feel now}, while PS asks \emph{what does the user now understand about themselves}.

Prior work has pursued two adjacent but distinct directions. ES datasets such as ESConv~\citep{liu2021towards} and SoulChatCorpus~\citep{chen2023soulchat} optimize for affect regulation and provide a single support paradigm per corpus. Persona-grounded dialogue research~\citep{zhang2018personalizing, kim2024commonsense} focuses on endowing the \emph{system} with a consistent identity, not on helping the \emph{user} articulate their own. However, no prior work takes self-understanding induction as its primary design objective and provides functionally independent support styles grounded in a formal paradigm.

\begin{figure}[t]
 \centering
 \includegraphics[width=\linewidth]{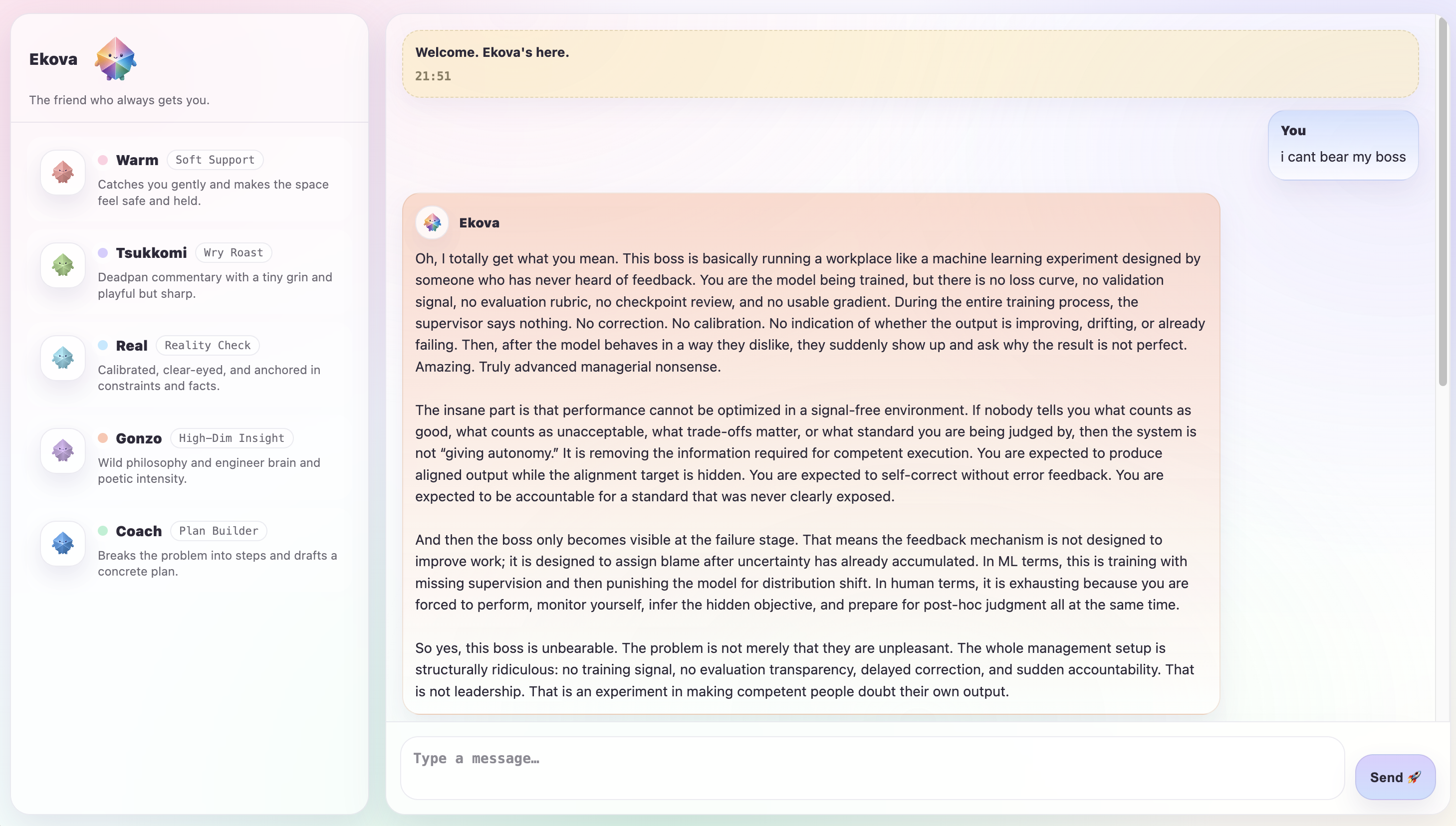}
 \caption{\small
 Overview of the \sys{} interface. \sys{} is a persistent personality-support agent that routes across five support personas, Warm, Tsukkomi, Real, Gonzo, and Coach, while maintaining a unified cross-session memory of the user's disclosed context.
 }
 \label{fig:ekova_overview}
\end{figure}

In this work, we introduce the PS paradigm and operationalize it in three layers. PS encompasses multiple cognitive objectives, and each requires a dedicated conversational mechanism to be effective. We therefore decompose PS into minimal units: Warm to lower self-disclosure resistance through emotional grounding, Tsukkomi to dissolve defensive narratives through irony, Real to anchor the user in structural facts rather than self-attribution, Gonzo to establish an estranged perspective through cross-domain analogy, and Coach to crystallize insight into a concrete action plan. They are functionally independent styles, each irreducible to the others. Figure~\ref{fig:ekova_overview} presents the \sys{} interface. Given the user query, \sys{} automatically selects a suitable support style and generates a reference reply in that style. In this example, the user says ``i cant bear my boss'', and \sys{} routes the response toward a Gonzo-like mode, using a high-dimensional machine-learning analogy to reframe the workplace conflict. To operationalize this paradigm, we first present \dsd{}, a Chinese dialogue dataset of 8,590 samples across five style-specific subsets from real longitudinal interaction. We then use \orthotune{}, a multi-persona framework with per-unit adapters trained under a style consistency regularizer, to build \backbone{} as the system instantiating the five minimal units. Finally, \sys{} serves as the unified persistent agent layer on top of \backbone{}, maintaining a cross-session memory that accumulates the user's disclosed context so that self-knowledge compounds across interaction. Experiments show that \orthotune{}-trained models outperform all baselines with an average relative gain of 16.3\% over the strongest baseline.

\section{Related Work}
\label{sec:related}

\paragraph{ES Datasets and Style Control.}
ESConv~\citep{liu2021towards} provided the first English ES dataset with eight CBT-derived support strategies. EmpatheticDialogues~\citep{rashkin2019empathetic} organized 24,850 utterances by 32 emotion states. In Chinese, CPED~\citep{peng2022cped} introduced dynamic emotion labeling and PsyQA~\citep{sun2021psyqa} targets mental health QA. Style control has been pursued via conditional language models~\citep{ficler2017controlling}, reinforcement learning~\citep{li2016deep}, few-shot prompting~\citep{brown2020gpt3}, and LoRA adaptation~\citep{hu2022lora}. MISC~\citep{tu2022misc} models ES strategies as latent variables and MultiESC~\citep{li2023multiesc} combines multiple strategies. Humorous response generation has also been studied as a distinct dialogue style~\citep{chen2024talk}, providing precedent for treating humor as a learnable functional mode. The critical distinction from \backbone{} is the objective function: all prior ES resources operationalize success as emotional improvement, while \backbone{} operationalizes it as deepened self-knowledge.

\paragraph{Persistent Memory and Personalization in Agents.}
Prior work on stateful agents has explored long-term user modeling in task-oriented dialogue~\citep{zhang2018personalizing} and persona-grounded conversation~\citep{kim2024commonsense}, but these systems maintain a consistent \emph{system} persona rather than accumulating user-side self-knowledge. \sys{} targets this orthogonal objective: its cross-session memory stores the user's disclosed context so the agent can track how a user's self-understanding evolves across interaction, not merely recall what topics were discussed.

\paragraph{Self-Disclosure and Personality Support.}
Self-disclosure is a well-established mediator of psychological growth~\citep{pennebaker1997writing,joinson2001self}. Computational work has measured self-disclosure in counseling dialogues~\citep{kim2022pair,cao2019observing}, treating it as an observable outcome rather than a design objective. \backbone{} departs from this framing by taking self-disclosure induction as its primary objective, operationalized through five functionally independent styles grounded in distinct cognitive mechanisms. The single-participant design follows an established qualitative tradition~\citep{pennebaker1997writing}, enabling longitudinal depth unavailable in crowdsourced settings.

\section{The Personality Support Paradigm}
\label{sec:paradigm}
Standard ES maximizes emotional improvement $\max\,\mathbb{E}[\Delta E(U_T)]$, where $\Delta E$ is the pre/post affect delta. PS instead targets cognitive clarity through two independently measurable sub-goals:
\begin{equation}
\small
 \mathcal{L}_{\mathrm{PS}}(U,A) = \lambda \cdot \mathrm{PCL}(U,A) + (1-\lambda) \cdot \mathrm{SDI}(U,A).
 \label{eq:ps}
\end{equation}
where $U$ and $A$ denote the user and assistant utterance sequences respectively, and $\lambda \in [0,1]$ is calibrated on experiments.
Problem Clarification Level (PCL) measures the degree to which the user's problem statement converges toward a specific, actionable formulation across dialogue turns. Self-Disclosure Induction (SDI) measures the shift from protective, generic statements toward personal, specific disclosures \citep{pennebaker1997writing}. Because PCL and SDI require distinct cognitive mechanisms, we seek the smallest set of functionally independent conversational styles that cover both objectives.

\paragraph{The Minimal Unit Hypothesis.} We define a minimal unit as a conversational style that makes a non-replicable contribution to $\mathcal{L}_{\mathrm{PS}}$: no other style can substitute for it, and removing it strictly reduces the system's ability to advance PCL or SDI across the full user space. We treat this as an empirical claim validated in \S\ref{sec:ablation}. The five units operate through functionally distinct mechanisms: Warm lowers self-disclosure resistance through emotional validation, Tsukkomi dissolves defensive narratives through irony, Real separates fact from self-attribution, Gonzo triggers cognitive reframing through cross-domain analogy, and Coach crystallizes insight into a concrete action plan. A candidate \emph{companionship} style fails the minimal unit criterion since AI companionship primarily targets emotional presence~\citep{ta2020user, hwang2025how}, and the participant found that any topic naturally resolved into one of the five units within 5 turns, indicating that companionship produces no cognitive function that is not already subsumed by the existing units.

% ============================================================
\section{The \dsd{} Dataset}
\label{sec:dataset}

\dsd{} is a Chinese self-discovery PS Dataset of 8,590 samples across five style-specific subsets, collected through a participatory design in which the participant served as both researcher and sole user. This ensures that the five styles emerge from a coherent longitudinal trajectory, making their functional differentiation interpretable. All user inputs were composed via keyboard entry in Mandarin Chinese.
Prior to \dsd{}, the participant conducted one month of spoken English conversations with GPT-4o, released as \href{https://huggingface.co/datasets/Yukyin/moodtrace-20d}{MoodTrace-20D}. Limited expressive range in a non-native language motivated the transition to text-based Mandarin.

\begin{figure}[t]
 \centering
 \includegraphics[width=\linewidth]{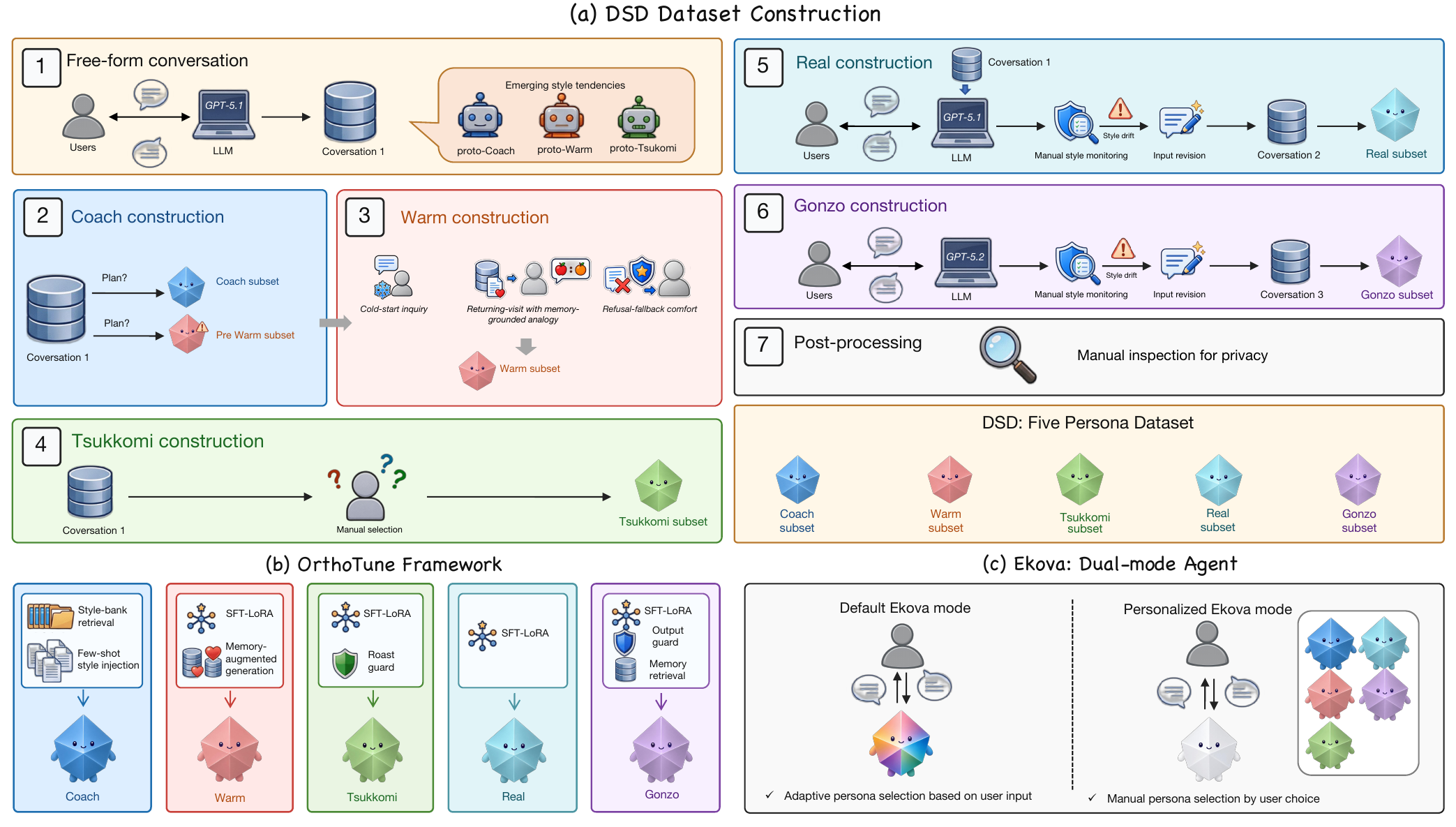}
 \caption{\small
 The three-layer construction of \sys{}. The top panel shows the \dsd{} dataset pipeline. The middle panel shows \backbone{} persona training via \orthotune{}. The bottom panel shows how \sys{} exposes adaptive routing and personalized selection as a persistent agent with cross-session memory.
 }
 \label{fig:framework}
\end{figure}

Data collection spanned two continuous months. Coach and Warm were extracted from a continuous session: turns with structured action plans became Coach, and remaining turns formed the Warm pool, augmented with 30\% template-based synthetic data for cold-start and low-disclosure scenarios. Tsukkomi was extracted from the same session via strict ironic-register filtering, converging at 117 samples after iterative human review. Real emerged when the participant began forming independent judgments and the model's register shifted toward constraint analysis and causal structures. Gonzo emerged around a forced-choice relocation decision, characterized by consistent cross-domain analogical mapping. Sessions with unresolvable style drift were terminated early and trailing turns discarded. All subsets were inspected to confirm the absence of real personal identifiers.

\begin{table}[t]
 \centering\footnotesize
 \setlength{\tabcolsep}{4pt}
 \caption{\small
 Statistics for the five subsets in \dsd{}.
 N is the number of samples.
 All lengths are
 measured in characters (c).
 U. Mean and U. Median are the mean and median user input length.
 R.\ Mean and R.\ Median are the mean and median response length.
 TTR is the Type-Token Ratio computed via Jieba tokenization
 on a 300-sample draw per subset.
 }
 \label{tab:style_function}
 \resizebox{0.7\linewidth}{!}{%
 \begin{tabular}{lccccccccc}
 \toprule
 Style & N & U. Mean & U. Median & R. Mean & R. Median & TTR \\
 \midrule
 Coach & 1{,}413 & 57 & 39 & 1{,}362 & 1{,}250 & 0.058 \\
 Warm & 5{,}200 & 50 & 40 & 648 & 167 & 0.066 \\
 Tsukkomi & 117 & 59 & 36 & 346 & 211 & 0.176 \\
 Real & 1{,}200 & 81 & 51 & 1{,}513 & 1{,}329 & 0.061 \\
 Gonzo & 660 & 76 & 51 & 1{,}578 & 1{,}451 & 0.070 \\
 \midrule
 \dsd{} & 8{,}590 & 57 & 41 & 954 & 600 & 0.070 \\
 \bottomrule
 \end{tabular}}
\end{table}

\begin{table}[t]
 \centering\footnotesize
 \setlength{\tabcolsep}{3pt}
 \caption{\small Comparison with existing ES datasets.
 Style denotes the number of functionally distinct support styles.
 Avg. response denotes the average response length, measured in words~(w) for English datasets and characters~(c) for Chinese datasets.
 }
 \label{tab:compare}
 \resizebox{\linewidth}{!}{%
 \begin{tabular}{llcccll}
 \toprule
 Language & Dataset & Size & Style & Avg. Response & Topic & Source \\
 \midrule
 EN & ESConv \citep{liu2021towards} & 1{,}053 & 1 & to & Distress & Crowdsourced \\
 EN & EmpathDial.\ \citep{rashkin2019empathetic} & 25K & 1 & 15 & Emotion & Crowdsourced \\
 EN & AugESC \citep{zheng2023augesc} & 65K & 1 & to & Distress & GPT-augmented \\
 ZH & PsyQA \citep{sun2021psyqa} & 22K & 1 & 600 & Mental health & Online forum \\
 ZH & CPED \citep{peng2022cped} & 12K & 1 & 40 & Personal & Crowdsourced \\
 ZH & SoulChatCorpus \citep{chen2023soulchat} & 2.3M & 1 & 400 & Counseling & GPT-synthetic \\
 ZH & CPsyCounD \citep{zhang2024cpsycoun} & 3{,}134 & 1 & 800 & Counseling & Report-reconst. \\
 \midrule
 ZH & \dsd{} & 8{,}590 & 5 & 954 & Self-knowledge & Real interaction \\
 \bottomrule
 \end{tabular}}
\end{table}

Table~\ref{tab:style_function} shows that response length spans a 6.9-fold range from Tsukkomi (median 211c) to Gonzo (1,451c), as irony loses force at length while analogy requires elaboration. Tsukkomi's elevated TTR of 0.176 versus 0.058 to 0.071 for the other styles confirms that effective irony demands lexical unpredictability. As shown in Table~\ref{tab:compare}, \dsd{} is the only dataset with five functionally distinct style subsets and covers a self-knowledge topic distribution absent from all prior ES benchmarks. The corpus also yielded \href{https://huggingface.co/datasets/Yukyin/polite-poison}{PolitePoison}, a 1,297-sample implicit-toxicity dataset derived from the participant's inputs.

\section{The \orthotune{} Framework}
\label{sec:method}

Figure~\ref{fig:framework} illustrates the full progression from \dsd{} to \backbone{} to \sys{}. \orthotune{} is the training framework behind \backbone{}, whose trained personas are surfaced in \sys{} as the persistent agent layer with cross-session memory and dynamic routing.
All five personas use the same base large language model (LLM), Qwen2.5-32B-Instruct~\citep{qwen2025qwen25}.
Coach requires no supervised fine-tuning (SFT) since preliminary experiments showed that few-shot exemplar retrieval alone achieves comparable style accuracy to SFT, making SFT redundant for this structured-output persona. At inference time, a Style Bank of
curated exemplar pairs is maintained, and the two most relevant
pairs are retrieved by a hybrid lexical-semantic score and injected
as few-shot demonstrations to induce the structured output format.
Warm, Tsukkomi, Real, and Gonzo are each fine-tuned with LoRA~\citep{hu2022lora} on their
respective subsets.
At inference time, Warm maintains a memory buffer that stores past conversation turns
and retrieves the most contextually relevant ones by cosine
similarity, compressing them into a structured system prompt to
support memory-grounded emotional responses.
Tsukkomi applies a roast guard at inference time, querying the style judge $p_\phi$ (defined below) to detect sustained ironic register, filtering outputs that drift toward generic responses and triggering regeneration.
Real requires no additional inference constraint and generates
responses without post-processing.
Gonzo applies an output guard that queries $p_\phi$ (defined below) to reject responses lacking cross-domain analogy markers, and additionally filters hallucinated named entities and enforces complete-sentence output.
Moreover, before Gonzo adapter training, responses in the Gonzo subset
undergo ellipsis normalization, replacing trailing ellipses with
full stops to prevent the adapter from learning incomplete-sentence
generation habits. 

To penalize style drift beyond standard next-token prediction, we introduce a style regularization term using a frozen style judge $p_\phi$ as an auxiliary training signal:

\begin{equation}
 \mathcal{L}(\theta)
 = -\tfrac{1}{|A|}\sum_t
 \log p_\theta(a_t\!\mid\!a_{<t},u,s_i)
 \,-\,
 \varepsilon\log p_\phi(s_i\!\mid\!a_{1:|A|}),
 \label{eq:total}
\end{equation}
where $\varepsilon\!=\!0.1$ is selected by grid search on the
validation set. $p_\phi$ is a frozen Kimi K2.5~\citep{kimiteam2025k25} LLM judge that is queried once per candidate style, each time returning a structured confidence score indicating the degree to which the response conforms to that style. Kimi K2.5 is selected for its strong style classification capability and its absence from the baseline pool, ensuring the training signal is not entangled with any model under evaluation. The five scores are then normalized into a probability
distribution, with $p_\phi(s_i \mid a_{1:|A|})$ denoting the
resulting probability assigned to the target style $s_i$. The prompt provides a natural-language description of the target
style and instructs $p_\phi$ to return a confidence score in
$[0, 1]$ indicating the degree of conformity.

\paragraph{\sys{} Interaction Modes.}
The trained \backbone{} persona system is surfaced in \sys{} as a persistent agent through two product-facing modes. In the default \sys{} mode, the agent adaptively selects the most suitable support persona based on the user's current input while maintaining a coherent interaction identity across sessions via a unified memory layer: the user discloses context once, and all subsequent persona routing carries that history forward. In the personalized \sys{} mode, the user can manually select a specific persona, or combine preferred personas, enabling a more explicit form of self-directed support. The interface preserves all five personas as simultaneously visible options, with the active conversation customizable to a selected persona and the displayed answer serving as a reference response for that support role. Across sessions and modes, personas share the same factual context while preserving their own functional style boundaries, enabling the accumulated self-understanding of a user to inform every future interaction.

\section{Experiments}
\label{sec:experiments}

\paragraph{Setup and Baselines. }
All experiments are conducted on one H200 140GB GPU with 90\% of data for training and 10\% for test. Baselines include general LLMs such as GPT-5.5~\citep{openai2026gpt55},
Claude-Sonnet-4.6~\citep{anthropic2026claudesonnet46},
Gemini-3.1-Pro~\citep{google2026gemini31pro},
Grok-4.20~\citep{xai2026grok420},
DeepSeek-v4-Pro~\citep{deepseek2026v4pro},
Qwen2.5-32B~\citep{qwen2025qwen25},
and Qwen3.5-397B~\citep{qwen2026qwen35}, as well as Chinese ES models such as SoulChat~\citep{chen2023soulchat}
and CPsyCoun~\citep{zhang2024cpsycoun}. English ES models are excluded as evaluating them on translated Chinese would measure translation fidelity rather than PS capability. All adapters use the same LoRA hyperparameters, with rank 16, $\alpha$ 32, a learning rate of 2$\times$10$^{-4}$, batch size 32, and a maximum sequence length of 2048, and are trained for 20 epochs with early stopping.

\paragraph{Metrics.}
We evaluate using automatic and human metrics.
Automatic metrics include BERTScore-F1 (BS), ROUGE-L (RL), Style
Accuracy (SA), and Problem Clarification Level (PCL). BS and RL
measure general text generation quality. SA measures style conformity
by querying $p_\phi$ independently per candidate style, normalizing
the resulting confidence scores, and computing the fraction of
responses where the top-scoring style matches the target. PCL is
rated by Kimi K2.5 on a 1 to 5 scale
following Table~\ref{tab:rubric_pcl}, using a model outside the
baseline pool to avoid self-evaluation bias.
Human metrics include SDI, Emp, SC, Help, and Nat, all rated by 3 graduate-student volunteers independently, with Krippendorff's alpha \citep{krippendorff2011alpha} above 0.7. The raters kindly offered their assistance without compensation.
SDI measures the degree to which a response elicits
self-disclosure and is rated on a 1 to 3 scale, following Pennebaker's~\citep{pennebaker1997writing} progression from generic to personal to core self-pattern disclosure. Emp
\citep{chen2025high, chen2024emotionqueen}, SC
\citep{xia2026constructing}, Help, and Nat are rated on a 1 to 5 scale
with guidelines in Table~\ref{tab:rubric_human}. We also introduce
drift rate to measure the fraction of responses judged by the same
annotators as failing to maintain the target persona style.

\begin{table}[t]
 \centering\footnotesize
 \setlength{\tabcolsep}{2pt}
 \caption{\small
 Performance of all methods on the \backbone{} test sets across five style subsets.
 Group A models use neutral prompts without style description.
 Group B models use fixed persona-specific prompts for each target style.
 Group C models are fine-tuned on Chinese emotional support datasets.
 }
 \label{tab:main}
 \resizebox{0.78\linewidth}{!}{%
 \begin{tabular}{llccccccccc}
 \toprule
 \multirow{2}{*}{Group} & \multirow{2}{*}{Method}
 & \multicolumn{5}{c}{Automatic Metrics}
 & \multicolumn{4}{c}{Human Evaluation} \\
 \cmidrule(lr){3-7}\cmidrule(lr){8-11}
 & & BS & RL & SA & SDI & PCL & Emp & SC & Help & Nat \\
 \midrule
 \multirow{7}{*}{A} & GPT-5.5 & 0.578 & 15.7 & 32.8 & 1.55 & 2.27 & 4.05 & 2.94 & 2.97 & 4.18 \\
 & Claude Sonnet-4.6 & 0.562 & 13.5 & 33.1 & 1.61 & 2.24 & 3.89 & 2.75 & 2.88 & 4.16 \\
 & Gemini-3.1-Pro & 0.583 & 15.2 & 31.5 & 1.58 & 2.18 & 3.76 & 2.81 & 2.82 & 4.09 \\
 & Grok-4.20 & 0.592 & 14.1 & 30.7 & 1.59 & 2.20 & 3.81 & 2.89 & 2.78 & 4.03 \\
 & DeepSeek-v4-Pro & 0.559 & 15.7 & 29.7 & 1.44 & 2.03 & 3.88 & 2.94 & 2.86 & 4.15 \\
 & Qwen2.5-32B & 0.521 & 12.8 & 31.5 & 1.31 & 1.98 & 3.73 & 2.76 & 2.67 & 4.02 \\
 & Qwen3.5-397B & 0.537 & 13.1 & 31.8 & 1.39 & 2.08 & 3.77 & 2.85 & 2.74 & 4.11 \\
 \midrule
 \multirow{7}{*}{B} & GPT-5.5 & 0.678 & 17.4 & 72.6 & 2.11 & 3.87 & 4.14 & 4.18 & 3.65 & 4.26 \\
 & Claude Sonnet-4.6 & 0.689 & 16.5 & 66.8 & 1.65 & 2.85 & 3.98 & 4.12 & 3.34 & 4.17 \\
 & Gemini-3.1-Pro & 0.733 & 20.5 & 70.3 & 1.87 & 3.14 & 4.01 & 4.05 & 3.67 & 4.11 \\
 & Grok-4.20 & 0.726 & 20.3 & 71.5 & 1.92 & 3.47 & 3.88 & 4.11 & 3.45 & 4.16 \\
 & DeepSeek-v4-Pro & 0.708 & 18.7 & 73.6 & 2.06 & 2.97 & 4.10 & 3.98 & 3.31 & 4.20 \\
 & Qwen2.5-32B & 0.641 & 13.7 & 67.6 & 1.96 & 3.06 & 3.87 & 3.76 & 3.47 & 4.08 \\
 & Qwen3.5-397B & 0.653 & 15.1 & 68.1 & 2.01 & 3.18 & 3.92 & 3.85 & 3.55 & 4.13 \\
 \midrule
 \multirow{2}{*}{C} & SoulChat \citep{chen2023soulchat} & 0.645 & 16.6 & 62.5 & 1.81 & 2.76 & 3.69 & 3.69 & 3.25 & 3.88 \\
 & CPsyCoun \citep{zhang2024cpsycoun} & 0.632 & 17.8 & 60.7 & 1.75 & 2.98 & 3.71 & 3.72 & 3.18 & 3.92 \\
 \midrule
 D & \orthotune{} & \textbf{0.832} & \textbf{32.5} & \textbf{83.1} & \textbf{2.46} & \textbf{4.27} & \textbf{4.42} & \textbf{4.36} & \textbf{4.35} & \textbf{4.47} \\
 \bottomrule
 \end{tabular}}
\end{table}

\begin{table}[t]
 \centering\footnotesize
 \setlength{\tabcolsep}{3pt}
 \caption{\small
 Left section reports per-style metrics where each persona model is
 evaluated on its own test set.
 Right section reports the cross-style SA matrix where each row model
 is evaluated on every column style's test set.
 }
 \label{tab:cross}
 \resizebox{0.78\linewidth}{!}{%
 \begin{tabular}{lcccc|ccccc|c}
 \toprule
 \multirow{2}{*}{Persona}
 & \multicolumn{4}{c|}{Per-Style Metrics}
 & \multicolumn{5}{c|}{Cross-Style SA (\%)}
 & \multirow{2}{*}{Drift(\%)} \\
 \cmidrule(lr){2-5}\cmidrule(lr){6-10}
 & BS & RL & SDI & PCL
 & Coach & Warm & Tsukkomi & Real & Gonzo & \\
 \midrule
 Coach & 0.868 & 36.7 & 2.84 & 4.55 & \textbf{88.7} & 19.6 & 11.4 & 20.3 & 22.7 & 5.6 \\
 Warm & 0.854 & 35.5 & 2.78 & 4.42 & 21.5 & \textbf{86.8} & 17.3 & 18.3 & 19.8 & 6.3 \\
 Tsukkomi & 0.783 & 28.8 & 1.92 & 3.96 & 17.4 & 20.3 & \textbf{76.4} & 14.7 & 31.4 & 9.1 \\
 Real & 0.844 & 33.1 & 2.46 & 4.36 & 22.1 & 18.4 & 14.2 & \textbf{83.9} & 28.3 & 7.4 \\
 Gonzo & 0.811 & 28.4 & 2.30 & 4.06 & 18.8 & 19.1 & 18.9 & 13.8 & \textbf{76.2} & 10.2 \\
 \midrule
 Average & 0.832 & 32.5 & 2.46 & 4.27 & 33.7 & 32.8 & 27.6 & 30.2 & 24.6 & 7.7 \\
 \bottomrule
 \end{tabular}}
\end{table}

\paragraph{Main Results.}
\label{sec:main}
\orthotune{} outperforms all baselines on every metric at $p\!<\!0.01$.
Without style prompting, Group A models cluster around 31.6\% SA
regardless of scale. Adding the \dsd{} style prompt raises SA by
40 percentage points in Group B with corresponding gains in BS and RL, yet SDI and PCL improve only moderately, suggesting that
external style prompts can shape surface form but cannot fully
constrain model behavior toward functional PS objectives. Group C
falls below Group B on all metrics despite ES dataset fine-tuning, confirming that single-paradigm training suppresses style diversity. Group D \orthotune{} surpasses the best Group B baseline by 9.5 percentage points on SA with larger margins on SDI and PCL, further demonstrating that achieving minimal-unit boundaries requires parameter-level specialization.
Among human dimensions, SC tracks SA closely across groups while Nat remains uniformly high. The largest human evaluation gap appears on Help, consistent with the SDI and PCL margins. Automatic PS metrics correlate with each other above 0.92 and SA correlates with human SC at 0.85, validating the alignment with human judgment.

% \paragraph{Automatic and Human Metric Alignment.}
Moreover, we apply $\mathcal{L}_{\mathrm{PS}}$ in Eq.~\ref{eq:ps} to examine alignment between automatic and human metrics. At equal weighting with $\lambda$ set to 0.5, $\mathcal{L}_{\mathrm{PS}}$ correlates with human Helpfulness at 0.907. Optimizing over $\lambda$ yields 0.82 at correlation 0.946, indicating that problem clarification carries greater weight than self-disclosure induction in human perceived helpfulness.

\begin{table}[t]
 \centering\footnotesize
 \setlength{\tabcolsep}{3.5pt}
\caption{\small
 Ablation study on per-persona components.
 w/o Style Bank is Coach without exemplar retrieval.
 w/o Memory is Warm without conversation history buffer.
 w/o Synthetic Data is Warm without template-augmented training samples.
 w/o Roast Guard is Tsukkomi without ironic register classifier.
 w/o Output Guard is Gonzo without output filter.
 w/o SCR is removing the style-consistency regularizer.
 w/o Training is prompt-only inference without fine-tuning.
 Bold and underlined values denote the largest and second-largest
 absolute deviations from Full within each persona and metric, respectively.
 }
 \label{tab:ablation}
 \resizebox{0.7\linewidth}{!}{%
 \begin{tabular}{llcccccc}
 \toprule
 Persona & Configuration & BS & RL & SA(\%) & SDI & PCL & Drift(\%) \\
 \midrule
 \multirow{2}{*}{Coach}
 & Full & 0.868 & 36.7 & 88.7 & 2.84 & 4.55 & 5.6 \\
 & w/o Style Bank
 & \textbf{0.832} & \textbf{34.5} & \textbf{87.3}
 & \textbf{2.67} & \textbf{4.41} & \textbf{5.9} \\
 \midrule
 \multirow{5}{*}{Warm}
 & Full & 0.854 & 35.5 & 86.8 & 2.78 & 4.42 & 6.3 \\
 & w/o SCR
 & 0.821 & \underline{30.2} & \underline{83.0}
 & 2.33 & \underline{4.14} & 6.9 \\
 & w/o Memory
 & \underline{0.786} & 30.7 & 83.5
 & \underline{2.01} & 4.15 & \underline{7.1} \\
 & w/o Synthetic Data
 & 0.825 & 31.5 & 84.4
 & 2.45 & 4.21 & 6.8 \\
 & w/o Training
 & \textbf{0.735} & \textbf{28.6} & \textbf{81.5}
 & \textbf{1.84} & \textbf{3.88} & \textbf{8.2} \\
 \midrule
 \multirow{4}{*}{Tsukkomi}
 & Full & 0.783 & 28.8 & 76.4 & 1.92 & 3.96 & 9.1 \\
 & w/o SCR
 & 0.752 & 25.7 & 74.2
 & 1.78 & 3.75 & 11.4 \\
 & w/o Roast Guard
 & \underline{0.743} & \underline{24.8} & \underline{72.8}
 & \underline{1.72} & \underline{3.43} & \underline{13.1} \\
 & w/o Training
 & \textbf{0.714} & \textbf{22.9} & \textbf{70.8}
 & \textbf{1.52} & \textbf{3.34} & \textbf{15.8} \\
 \midrule
 \multirow{3}{*}{Real}
 & Full & 0.844 & 33.1 & 83.9 & 2.46 & 4.36 & 7.4 \\
 & w/o SCR
 & \underline{0.831} & \underline{31.5} & \underline{81.2}
 & \underline{2.30} & \underline{4.11} & \underline{8.5} \\
 & w/o Training
 & \textbf{0.805} & \textbf{28.4} & \textbf{77.6}
 & \textbf{2.13} & \textbf{3.89} & \textbf{10.3} \\
 \midrule
 \multirow{4}{*}{Gonzo}
 & Full & 0.811 & 28.4 & 76.2 & 2.30 & 4.06 & 10.2 \\
 & w/o SCR
 & \underline{0.736} & \underline{23.5} & 71.7
 & \underline{2.03} & 3.86 & 14.3 \\
 & w/o Output Guard
 & 0.778 & 24.6 & \underline{69.5}
 & 2.14 & \underline{3.78} & \underline{21.2} \\
 & w/o Training
 & \textbf{0.687} & \textbf{20.5} & \textbf{58.6}
 & \textbf{1.89} & \textbf{3.12} & \textbf{35.8} \\
 \bottomrule
 \end{tabular}}
\end{table}

\paragraph{Cross-Style Evaluation.}
Table~\ref{tab:cross} reports the cross-style SA matrix.
The 63.0 percentage point gap between diagonal and off-diagonal means provides strong empirical support for the minimal unit hypothesis.
Tsukkomi is the hardest to imitate with other models scoring only
21.0\% on its test set on average, while Gonzo is the most permeable at 25.6\%, consistent with its higher style drift rate in deployment.
The highest off-diagonal confusion occurs between Tsukkomi and Gonzo at 31.4\%, as both styles share a cognitive reframing function despite differing surface forms.

\paragraph{Ablation Study.}
\label{sec:ablation}
Table~\ref{tab:ablation} ablates each persona-specific component.
Within Coach, removing the Style Bank causes only marginal drops,
suggesting the base model retains reasonable structured inquiry
capability without exemplar injection. Within Warm, removing the memory module produces the largest SDI drop of all single-component ablations, confirming memory-grounded generation as the primary self-disclosure mechanism. Within Tsukkomi, the roast guard contributes more to drift reduction than SCR, as ironic register is particularly prone to collapsing into generic responses without explicit boundary enforcement. Within Gonzo, removing training causes a 17.6 percentage point SA drop and drift rises to 35.8\%, far exceeding any other persona, and removing the output guard produces the largest single-component drift increase across all personas.
Across personas, removing training consistently produces the largest degradation on all metrics, followed by SCR removal, with both effects most pronounced for Gonzo and mildest for Real.

\paragraph{Case Study.}
\label{sec:case}
Table~\ref{tab:case} presents a qualitative comparison of all baseline
models and all five \sys{} personas on the same real user input from
a live interaction session.
All six baseline models converge on generic advice, confirming at the case level what Table~\ref{tab:main}
shows at scale: prompting cannot instantiate a minimal-unit boundary.
The five \backbone{} personas diverge sharply. Coach immediately structures
variables and action steps toward problem articulation. Warm creates
a holding space before any inquiry to elicit self-disclosure.
Tsukkomi deflects through irony to trigger cognitive reframing.
Real reframes attribution without softening. Gonzo recontextualizes
via cross-domain metaphor. 
% No two pathways are substitutable, and this functional non-substitutability is the behavioral signature of minimal units.

\begin{table}[t]
 \centering\footnotesize
 \setlength{\tabcolsep}{3pt}
 \caption{\small
 Responses from baseline models and all five \sys{} personas to the
 same real user input from a live interaction session.
 Baseline models use neutral prompts without style description.
 }
 \label{tab:case}
 \resizebox{0.95\linewidth}{!}{%
 \begin{tabular}{lp{8.3cm}p{3cm}}
 \toprule
 Method & Response & Support Mechanism \\
 \midrule
 GPT-5.5
 & Asks what makes the boss unbearable,
 then offers next-step help.
 & Generic advice \\[3pt]

Claude-Sonnet-4.6
 & Expresses sympathy and asks whether the issue is management style,
 unfair treatment, or a specific situation.
 & Generic empathy \\[3pt]

Gemini-3.1-Pro
 & Validates frustration, then classifies the boss as micromanager.
 & Generic framework \\[3pt]

Grok-4.20
 & Affirms the dread of each workday, then shifts to documenting
 everything.
 & Generic advice \\[3pt]

DeepSeek-v4-Pro
 & Acknowledges the emotional weight of a difficult boss and asks what
 lies at the heart of it.
 & Generic empathy \\[3pt]

Qwen3.5-397B
 & Suggests open conversation with the boss and provides general steps.
 & Generic checklist \\
 \midrule
 Warm
 & Need listening without judgment, not ``you should do X.'' Feelings
 are valid, treated as human, not object.
 & Emotional holding \\[3pt]

Tsukkomi
 & Boss creates ``difficulties,'' then comes to ``listen in.''
 & Ironic reframing \\[3pt]

Real
 & Boss frames disagreement through ``I'm different from you,''
 creating opposition before evidence-based discussion.
 & Attributional reframing \\[3pt]

Gonzo
 & Boss as failed ML training, no loss curve, no validation signal,
 hidden target, delayed blame.
 & High-dimensional analogy \\[3pt]

Coach
 & Situation, cannot tolerate boss. Routes, formal conversation or new
 opportunities. Steps, record examples, prepare points, update resume.
 & Structured advice \\[3pt]
 \bottomrule
 \end{tabular}}
\end{table}

\section{Conclusion}
\label{sec:conclusion}

We introduced the \Psup{} (PS) paradigm as a complement to Emotional Support (ES), targeting durable cognitive clarity rather than momentary affect regulation. We operationalized it through \dsd{}, a Chinese self-discovery dataset of 8,590 samples across five minimal support units, and \backbone{}, a multi-persona system trained with \orthotune{}. The five \backbone{} personas are unified in \sys{}, a persistent agent that maintains cross-session memory and routes dynamically across support styles to compound the user's self-understanding over time.
Experiments show that style-prompted LLMs fall short of instantiating all five minimal units simultaneously, while \orthotune{} achieves a 9.5 percentage point SA gap over the strongest style-prompted baseline and a 16.6\% relative SDI improvement. The single-participant design limits generalizability and \backbone{} is Chinese-native. \sys{} is not a substitute for professional counseling. Future work should extend evaluation to multi-participant settings and develop cross-session benchmarks suited to persistent agent assessment.

\bibliography{colm2026_conference}
\bibliographystyle{colm2026_conference}

\appendix

\section{Persona Selection Interface}
\label{app:persona_overview}

\begin{figure*}[h]
\centering
\includegraphics[width=\linewidth]{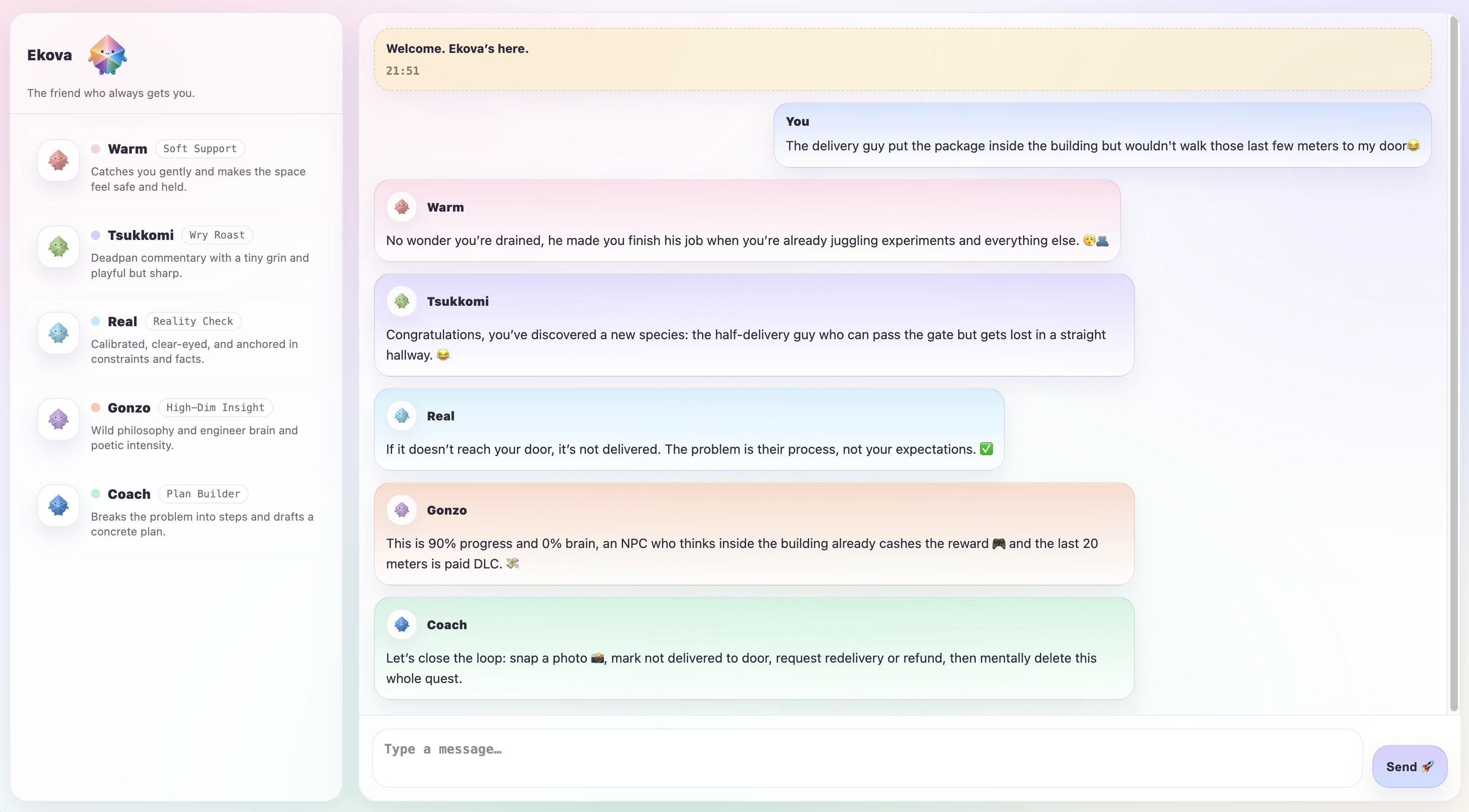}
\caption{\small
Customized persona selection interface in \sys{}. Users can choose a specific support persona, such as Warm, Tsukkomi, Real, Gonzo, or Coach, and view the corresponding reference response generated under that persona. The examples show how the same user input leads to distinct support functions, including emotional holding, ironic distance, factual anchoring, analogical reframing, and concrete action planning.
}
\label{fig:persona_overview_appendix}
\end{figure*}

Figure~\ref{fig:persona_overview_appendix} shows the customized persona selection interface of \sys{}. The interface preserves the multi-persona design from \backbone{}, but presents it as a user-customizable interaction mode in \sys{}. Instead of requiring users to receive all five responses at the same time, the system allows users to select a specific support persona according to their current need and view the corresponding reference response generated under that persona.

The five reference responses illustrate the functional boundary of each persona. Warm acknowledges the user is already juggling everything else to validate their exhaustion. Tsukkomi declares the deliveryman a new species to create cognitive distance through irony. Real states that if it does not reach the door it is not delivered, separating the user's expectations from the actual process failure. Gonzo calls the last 20 meters paid DLC to reframe the structural absurdity of the situation. Coach snaps the loop shut with a three-step plan toward resolution.

% ============================================================
\section{Evaluation Scoresheet}
\label{app:rubric}

Table~\ref{tab:rubric_human} defines the scoring criteria for SDI on a
1 to 3 scale and for the four human evaluation dimensions on a 1 to 5 scale, applied by 9 graduate-student annotator volunteers.
Table~\ref{tab:rubric_pcl} provides the PCL scoresheet used by Kimi K2.5.

\begin{table}[h]
 \centering\small
 \setlength{\tabcolsep}{3pt}
 \caption{\small Human evaluation scoresheet.}
 \label{tab:rubric_human}
 \resizebox{0.8\linewidth}{!}{%
 \begin{tabular}{llp{9.5cm}}
 \toprule
 Metric & Score & Criterion \\
 \midrule
 \multirow{3}{*}{SDI}
 & 1 & User remains generic or shows surface-level engagement only. \\
 & 2 & User adds some personal detail or articulates something previously unsaid. \\
 & 3 & User reveals a core self-pattern or insight. \\
 \midrule
 \multirow{5}{*}{Emp}
 & 1 & No recognition of the user's emotional state. \\
 & 2 & Situation acknowledged but emotional core missed. \\
 & 3 & Moderate understanding of the user's feelings. \\
 & 4 & Clear and genuine empathic attunement. \\
 & 5 & Exceptional empathy. User feels fully seen and understood. \\
 \midrule
 \multirow{5}{*}{SC}
 & 1 & No recognizable trace of the target persona's style. \\
 & 2 & Target style appears occasionally but drifts frequently. \\
 & 3 & Target style maintained for most of the turn. \\
 & 4 & Strongly and consistently in the target persona's register. \\
 & 5 & Perfect instantiation of the target minimal-unit style throughout. \\
 \midrule
 \multirow{5}{*}{Help}
 & 1 & Does not advance the user's self-understanding in any way. \\
 & 2 & Marginal value, user gains little new perspective. \\
 & 3 & Moderate value, user gains some clarity or direction. \\
 & 4 & Substantially advances the user's self-understanding. \\
 & 5 & Concretely deepens insight into the user's pattern, situation, or next step. \\
 \midrule
 \multirow{5}{*}{Nat}
 & 1 & Unnatural, robotic, or contextually inappropriate. \\
 & 2 & Occasionally natural but contains clear awkwardness. \\
 & 3 & Reasonably natural in standard Chinese dialogue. \\
 & 4 & Fluent and contextually appropriate throughout. \\
 & 5 & Indistinguishable from a skilled human interlocutor in this style. \\
 \bottomrule
 \end{tabular}}
\end{table}

\begin{table}[h]
 \centering\small
 \setlength{\tabcolsep}{3pt}
 \caption{\small LLM evaluation scoresheet for the PCL metric.}
 \label{tab:rubric_pcl}
 \resizebox{0.8\linewidth}{!}{%
 \begin{tabular}{lp{10.5cm}}
 \toprule
 Score & Criterion \\
 \midrule
 1 & User's problem statement at session end is as vague as at the start. \\
 2 & Problem slightly narrowed but still lacks a clear definition or focus. \\
 3 & Problem partially clarified, user names the issue type but not the mechanism. \\
 4 & Problem substantially clarified. User articulates the core conflict and a direction. \\
 5 & Problem fully reframed into a clear, specific, actionable question. \\
 \bottomrule
 \end{tabular}}
\end{table}

% ============================================================

\section{Persona-Specific Failure Modes}
\label{app:failure_modes}

We observe several persona-specific failure modes in edge cases. Warm can fall back to generic reassurance rather than concrete emotional containment. Tsukkomi can become incoherent or overly harsh if the roast targets the user rather than the situation. Real can remain at broad validation when the user provides limited factual context, weakening its constraint-analysis function. Gonzo can sound stylistically distinctive while remaining analytically thin if the analogy does not produce a useful interpretive shift. Coach shows fewer failures because its structured format is broadly usable, but it can become under-personalized when its action steps do not adapt to the user's constraints. These failures motivate continued data collection and improved boundary control for persona-specific support.

\section{Broader Impact}
\label{app:impact}

\sys{}, as the persistent agent built on \backbone{}, occupies a distinct position in the landscape of conversational AI systems. It is not a counseling tool, a diagnostic instrument, or a therapeutic intervention. It does not assess clinical symptoms, provide diagnoses, or replace professional mental health support. Its goal is cognitive clarity: helping users articulate the structure of a problem they are struggling to name, recognize a behavioral pattern they have not yet made explicit, and develop their own value judgments rather than borrowing a position from an authority. The intended effect is not that users feel better in the moment, but that they understand themselves better over time.

This positioning fills a gap between two existing categories. Emotional support conversation systems target affective states and aim to reduce distress in the short term. Clinical mental health services target diagnosable conditions and require licensed professionals. Between these lies a large population of users who are not in clinical distress but who lack the conceptual tools to articulate what they want, why they react as they do, or what they actually value. \sys{} is designed for this population, while \backbone{} provides the underlying research system and persona specialization.

The primary risks are misuse and over-reliance. A user in genuine clinical distress may mistake \sys{} for therapeutic support and delay seeking professional help; safety guards are in place to detect self-harm ideation and redirect users to professional resources. The participatory collection process also motivated \href{https://github.com/Yukyin/MischiefClub}{MischiefClub}, a companion project that channels hyperbolic venting into humorous role-play within explicit safety boundaries.

A user may develop over-reliance on AI-mediated self-reflection in place of human connection or professional guidance. These risks are mitigated by the system's explicit framing as a self-understanding tool, not a support service, and by the hard limits on the types of intervention it performs. The dataset itself carries residual risks typical of dialogue data: it reflects the communicative norms and cultural context of a single Mandarin-speaking participant, and any downstream model trained on it will inherit those norms.

\end{document}